\documentclass[journal]{IEEEtran}

\usepackage{subfigure}
\usepackage{url}
\usepackage{amsmath}
\usepackage{mathptmx}
\usepackage{mathtools}
\usepackage{multirow}
\usepackage{multicol}
\usepackage{changepage}
\usepackage{algorithm}
\usepackage{algcompatible}
\usepackage{balance}
\usepackage{comment}
\usepackage{tabularx}
\usepackage{tabulary}
\usepackage{footnote}
\usepackage{multicol}
\usepackage{xcolor}

\usepackage{graphicx}

\usepackage[all]{nowidow}
\begin{document}
\bibliographystyle{plain} 


\title{Three-dimensional Backbone Network for 3D Object Detection in Traffic Scenes}
\author{Xuesong Li, Jose Guivant,~\IEEEmembership{Member,~IEEE,} Ngaiming Kwok, Yongzhi Xu, Ruowei Li, Hongkun Wu$^{*}$

\thanks{X. Li (e-mail: xuesong.li@unsw.edu.au); J. Guivant (e-mail: j.guivant@unsw.edu.au); N. Kwok (e-mail: nmkwok@unsw.edu.au); R. Li (e-mail: ruowei.li@unsw.edu.au); H. Wu (e-mail: hongkun.wu@unsw.edu.au). All above authors are with the School of Mechanical Engineering, University of New South Wales, Sydney, NSW 2052, AU} 
\thanks{Y. Xu is with the school of Civil and Environmental Engineering, the University of New South Wales, Sydney, NSW 2052, AU (e-mail: y.xu@unsw.edu.au).} 
\thanks{
Corresponding Author(*): H. Wu (e-mail: hongkun.wu@unsw.edu.au).}}

\maketitle

\begin{abstract}
The task of detecting 3D objects in traffic scenes has a pivotal role in many real-world applications. However, the performance of 3D object detection is lower than that of 2D object detection due to the lack of powerful 3D feature extraction methods. To address this issue, this study proposes a 3D backbone network to acquire comprehensive 3D feature maps for 3D object detection. It primarily consists of sparse 3D convolutional neural network operations in point cloud. The 3D backbone network can inherently learn 3D features from the raw data without compressing the point cloud into multiple 2D images. The sparse 3D convolutional neural network takes full advantage of the sparsity in the 3D point cloud to accelerate computation and save memory, which makes the 3D backbone network feasible in real-world application. Empirical experiments were conducted on the KITTI benchmark and comparable results were obtained with respect to the state-of-the-art performance.
\end{abstract}


\IEEEpeerreviewmaketitle

\section{\uppercase{Introduction}}\label{sec:introduction}

\begin{figure}[!tbp]
	\includegraphics[width = \linewidth]{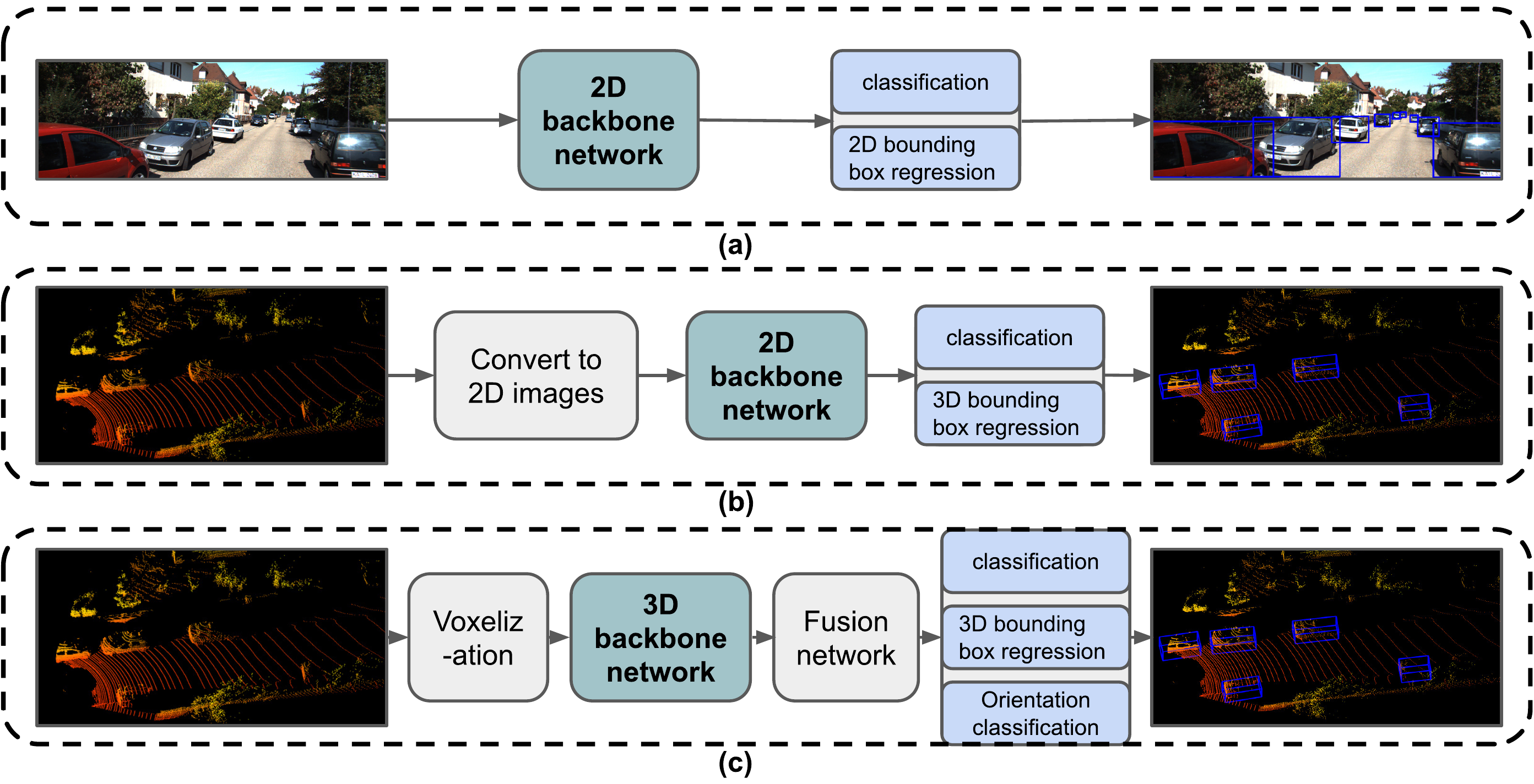} 
	\caption{Network structure for object detection. {\bf (a)} is the common framework for 2D object detection; {\bf (b)} is the general framework for 3D object detection; {\bf (c)} is our proposed framework.}
\label{fig:framework}
\end{figure}

\IEEEPARstart{T}{hree-dimensional} (3D) object detection in traffic scenes can provide accurate 3D spatial location of targets and their geometric shapes. This capability is vital for various traffic applications, such as autonomous driving vehicles and delivery robots. Over the past ten years, a number of backbone networks based on two-dimensional (2D) Convolutional Neural Network (CNN), such as AlexNet \cite{Krizhevsky:2012:ICD:2999134.2999257}, VGG \cite{VGG_network}, and ResNet \cite{he2016deep} were proposed to extract features for computer vision tasks [see Fig. \ref{fig:framework}(a)], and those have made a remarkable contribution to the development of 2D object detection research \cite{Ren_2017_CVPR, visapp19}. On the other hand, the CNN based backbone network has not been fully explored for 3D object detection. The main reason is that the 3D data often has irregular structure, for which the CNN can not be directly applied. With the additional dimension, the amount of processing needed in performing CNN increases dramatically to an extent beyond the processing capability of common hardware. This work seeks to investigate how 3D CNN can be efficiently applied for the object detection in an improved sparse CNN \cite{SubmanifoldSparseConvNet,DBLP:journals/corr/Graham15}. 

3D object detection based on point cloud has two main challenges: irregular data structure and expensive computation costs in the 3D CNN. To address these issues, the irregular data is converted into a regular structure, such as bird view \cite{DBLP:journals/corr/LiZX16}, front view \cite{DBLP:journals/corr/ChenMWLX16}, or 3D voxel grids \cite{DBLP:journals/corr/abs-1711-06396}. Then, efficient and memory-friendly feature extractors, such as a 2D CNN based network and a shallow 3D CNN, are applied for detection, as shown in Fig. \ref{fig:framework}(b). A multi-view 3D (MV3D) \cite{DBLP:journals/corr/ChenMWLX16} generates 3D proposals by running a 2D CNN on projected bird view images. 

VoxelNet \cite{DBLP:journals/corr/LiZX16} builds three layers of 3D CNN to extract 3D features for region proposals. However, these 3D data processing methods, such as compressing 3D data into 2D or shallow 3D layers, are merely concentrated on partial 3D features for detection, consequently failing to achieve good performance in 3D detection. Given that 2D backbone networks have demonstrated a considerable success in object detection from 2D data sets, it is presumable that the 3D backbone network would also yield a similar performance towards object detection in a 3D data set. However, according to our knowledge, there is no available 3D feature extraction backbone network for 3D object detection. Therefore, this study proposes a 3D backbone network to extract better 3D features for object detection, as shown in Fig. \ref{fig:framework}(c). 

The 3D backbone network presented here is made up primarily of the sparse 3D CNN \cite{SubmanifoldSparseConvNet,DBLP:journals/corr/Graham15,3DSemanticSegmentationWithSubmanifoldSparseConvNet}. Sparse 3D CNN takes advantage of spatial sparsity in the point cloud data, which can significantly reduce memory and processing costs. The proposed backbone network includes two different architectures: 3DBN-1 and 3DBN-2. The 3DBN-1 has only one forward path from bottom to top, which generates 3D feature maps with various resolutions and abstractions. The forward network, 3DBN-2, incorporates one extra backward path from top to bottom, which propagates the high semantic but low-resolution feature map back to the previous feature map in a low semantic level. They are further fused to obtain high-resolution and high semantic feature maps. 

The whole detection scheme is illustrated in Fig. \ref{fig:framework}(c). The 3D backbone network must process data with a regular format, such as a 3D voxel representation. Therefore, before being fed into the 3D backbone, the raw point cloud is voxelized as homogeneous 3D grids. After 3D feature maps are produced by the backbone network, the fusion network is used to dilate the sparsity of the feature maps to generate 3D proposals. For 3D object detection, there are two basic tasks, classification and bounding box regression. In addition, we embed one extra classification to find the heading of the predicted boxes. Opposite directions are considered to be identical in the bounding box regression task and different in the extra heading classification task, which will not lead to discontinuities in the heading offsets of predicted boxes with respect to other anchors. 

We evaluated the proposed method with the KITTI benchmark \cite{geiger2012we} for 3D object detection. Experimental results show that the proposed 3D backbone network can extract 3D features for object detection without significantly increasing the computation load. Its detection accuracy also outperforms other well-known 3D object detection methods, such as VoxelNet\cite{ DBLP:journals/corr/abs-1711-06396}, MV3D\cite{DBLP:journals/corr/ChenMWLX16}, AVOD-FPN \cite{avod_3d}, and F-PointNet\cite{DBLP:journals/corr/abs-1711-08488}. The source code has been made publicly available\footnote[1]{The source code can be found: \url{https://github.com/Benzlxs/tDBN}}.

The contributions of this work are:
\begin{enumerate}
    \item proposing a 3D backbone network for 3D object detection;
    \item designing a 3D detection network by using a fusion network;
    \item developing a new method to predict the orientation of the object.
\end{enumerate}

The rest of the paper is organized as follows. Section \ref{sec:Related_work} introduces the related work, followed by Section \ref{sec:feature_pyramid_network}, which illustrates how to build the 3D backbone network with sparse CNN. Additional procedures about 3D object detection are presented in Section \ref{sec:3D_object_detection_network}. Experiments of the proposed method are described in Section \ref{sec:experiments}. Section \ref{sec:conclusion} concludes our work and summarizes the contributions.

\section{\uppercase{Related work}}\label{sec:Related_work}

In the past serveral years, many 3D detection frameworks were proposed. Those could be roughly divided into three categories. The first one is the image-based detection, which only relies on 2D imagery to predict 3D bounding boxes. The second type is 3D point cloud based detection, which takes 3D data as input. The last category of work hybridizes the former two strategies. 

\subsection{Image-based 3D detection}
Given recent remarkable achievements yielded by 2D object detection from an image, a variety of 3D detectors \cite{mono_3d_objects, DBLP:journals/corr/MousavianAFK16, DBLP:journals/corr/ChabotCRTC17, Xiang2015Datadriven3V} have been proposed by integrating well-developed 2D object detectors and prior information of target objects, such as shape, context, and occlusion pattern to infer 3D objects from a monocular image. The 3D object proposal method \cite{mono_3d_objects} generates 3D proposals in 3D layouts with context priors, and then projects 3D proposals back into regions of interest in the image to crop corresponding convolutional features for further classification and bounding boxes regression. Mousavian et al. \cite{DBLP:journals/corr/MousavianAFK16} estimated 3D object poses by using geometric transformation constraints imposed by perspective projection from 3D bounding boxes to 2D windows in the image. 

Deep MANTA \cite{DBLP:journals/corr/ChabotCRTC17} contains a coarse-to-fine object proposal framework and enables the final refining network for multitasks: 2D object detection, part localization, visibility characterization, and 3D template estimation. Since projecting a 3D scene onto a 2D plane is an irreversible process, relevant 3D information is lost in forming the 2D image. Thus, 3D object detector based on monocular image often suffers from low detection accuracy. Therefore, an object detector that can manipulate 3D point clouds rather than being confined to 2D data is preferred to avoid information loss during the detection.

\subsection{3D point cloud-based detection}
The 3D point cloud can directly represent real-world scenes and is regarded as an important information source to detect objects. The common procedure for processing the point cloud is converting an irregular point cloud into a regular data representation, such as 3D voxels or 2D view images, followed by the 2D CNN to extract features. The work reported in \cite{DBLP:journals/corr/LiZX16} encoded the point cloud into a cylindric-like image in which every valid element carries 2-channel data. The approach \cite{DBLP:journals/corr/ChenMWLX16} includes a new data representation and converts the 3D data into bird view representation, including multiple height maps, one density map, and one intensity map. It then applies a 2D CNN on these bird view images to propose 3D object candidates. These methods use the mature 2D CNN framework to perform detection efficiently, but fail to learn the features in 3D space. Rather than compressing 3D data into 2D planes, other methods \cite{DBLP:journals/corr/Li16p, 5439956} discretize the point cloud into 3D voxel grids and process them with 3D convolution. 

VoxelNet \cite{DBLP:journals/corr/abs-1711-06396} and SECOND \cite{s18103337} adopt PointNet \cite{DBLP:journals/corr/QiSMG16, qi2017pointnetplusplus} to extract features from the low resolution raw point cloud for each voxel. It then employs a shallow 3D CNN network to convert 3D feature maps into 2D feature maps. A powerful 2D-CNN-based network can then be used to further extract features and generate proposals. These methods learn object features in the original 3D space without dimensional reduction. However, the 3D CNN layers are not deep enough to extract abstract 3D features and they still heavily depend on the 2D region proposal network. Therefore, this study proposes a deeper 3D backbone network to fully exploit feature learning in 3D feature space rather than merely using a 2D CNN backbone network.

\begin{figure*}[!h]
 \begin{adjustwidth}{0 cm}{}
	\includegraphics[width=\textwidth]{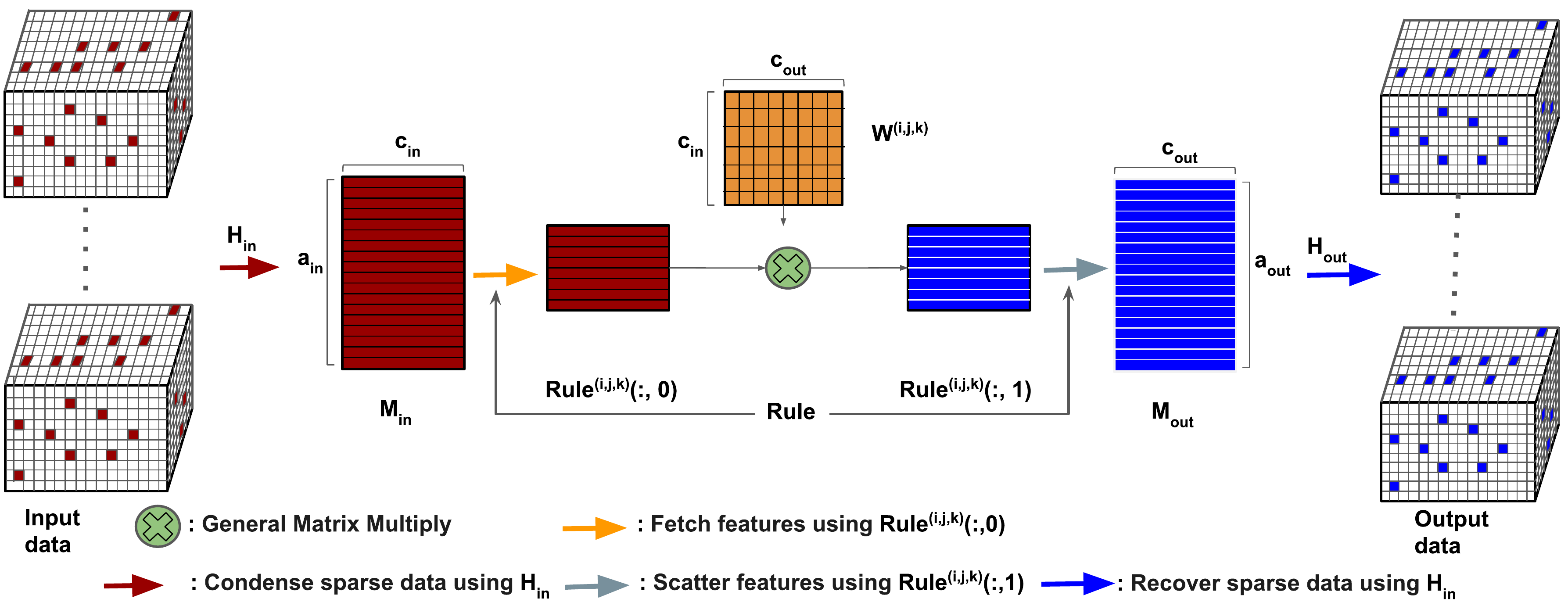} 
 \end{adjustwidth}	
	\caption{Data flow pipeline from input to output by the use of a sparse CNN. The input data is firstly converted into the input feature matrix ${\bf M}_{in}$ by using the hash table ${\bf H}_{in}$, the Rule book is built to fetch data from ${\bf M}_{in}$ for convolutional operations with the corresponding convolutional weighting matrix ${\bf W}^{i,j,k}$, and the output feature matrix ${\bf M}_{out}$ is obtained. Finally, the output feature matrix can be converted back to the output data through output hash table ${\bf H}_{out}$.}
\label{fig:sparseCNN}
\end{figure*}

\subsection{Detection by fusing image and 3D point cloud}
A number of hybrid methods have been adopted for fusing the image and 3D point cloud for object detection \cite{DBLP:journals/corr/abs-1711-08488, DBLP:journals/corr/ChenMWLX16, liang2019multi, wang2019frustum}. The 3D proposals are usually generated using the RGB image or the point cloud only, then the feature maps extracted from both information sources are concatenated for further inference. The AVOD-FPN \cite{avod_3d} predicts bounding box proposals on the LiDAR data with a conventional CNN, and a region-based fusion network that merges features from two 2D images: LiDAR bird view and RGB image for every proposal. 

Frustum PointNets \cite{DBLP:journals/corr/abs-1711-08488, wang2019frustum} employ a pre-trained 2D object detector to detect objects in the image and extrudes corresponding 3D bounding frustums in point clouds. PointNet \cite{DBLP:journals/corr/QiSMG16} is then employed to process point cloud within every frustum to detect 3D objects. However, these approaches are very sensitive to synchronization error between two sensors: camera and 3D Lidar and also complicate the sensing system. In contrast, we introduce an approach which is fully based on 3D imagery and mitigate limitations of those mentioned approaches, in particular, those related to accuracy and processing cost.

\section{3D Backbone Network}\label{sec:feature_pyramid_network}

The proposed 3D backbone network stems from a sparse CNN \cite{SubmanifoldSparseConvNet, DBLP:journals/corr/Graham15}, and its working principle is first investigated in this section, followed by a detailed description of the proposed method.

\subsection{Sparse CNN}

\begin{figure*}[!h]
	\includegraphics[width=\textwidth]{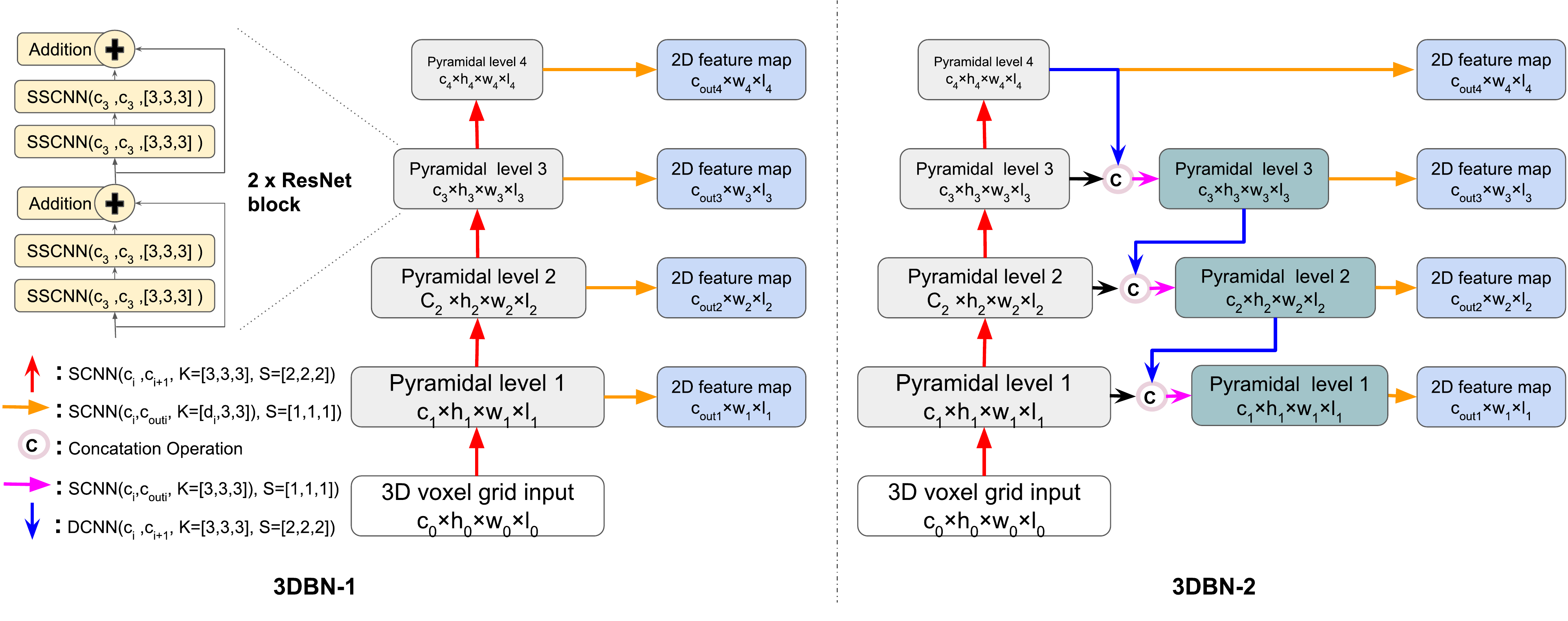} 
	\caption{3D Backbone Network: SSCNN denotes the submanifold sparse 3D CNN, SCNN is the sparse 3D CNN, DCNN is the sparse 3D deconvolutional neural network, and $K$ and $S$ are the convolutional kernel and stride, respectively.}
\label{fig:3DBN}
\end{figure*}

The notation and glossary are introduced first. The lowercase, capital, and bold capital are used for representing the scalar, vector, and matrices, respectively. The input data is referred to as the data fed into the current layer and it can be the raw data in the input layer, or feature maps in the hidden layer. The output data is generated by the current layer which can be the feature maps in a hidden layer or an output layer. We assume that the size of input data is $c_{in} \times h_{in} \times w_{in} \times l_{in}$ where the 3D input field size is $h_{in}\times w_{in} \times l_{in}$, and $c_{in}$ is the number of 3D feature map channels. The size of 3D CNN filter is ($k \times k \times k \times c_{in} \times c_{out}$, $s$) in which $k$ is the convolutional kernel size, $s$ is the stride size, and $c_{out}$ is the number of output channels. If this filter is used to process the aforementioned input data, the size of the output data will be $c_{out} \times h_{out} \times w_{out} \times l_{out}$, where $h_{out} = (h_{in}-k)/s$, $w_{out} = (w_{in}-k)/s$, and $l_{out} = (l_{in}-k)/s$. 

The conventional 3D CNN is computation demanding and has high memory consumption; therefore, it is not widely used in processing 3D data. In real-world applications, most 3D data, such as point clouds, are very sparse. After being voxelized, the 3D input data is assumed to enclose $h_{in}\times w_{in} \times l_{in}$ 3D grids, in which only $a_{in}$ contain points or active points\footnote[2]{The point in the current layer is active if it satisfies one of the following criteria: 1) its vector magnitude in the input layer is not zero; 2) in the hidden layer, it receives information from the non-zero vector input layer.}, usually, $a_{in} \ll h_{in}\times w_{in} \times l_{in}$, thus, the majority of memory and computation is spent on empty data areas. By taking full advantage of data sparseness, the 3D sparse CNN can avoid the computation and memory cost of empty points in the input data. The pipeline of sparse CNN is shown in Fig. \ref{fig:sparseCNN}.

The input data, a 4D tensor with the size: $c_{in} \times h_{in} \times w_{in} \times l_{in}$, is first encoded into the input feature matrix $\textit{\textbf{M}}_{in}$ with the hash table $\textit{\textbf{H}}_{in}$. The resultant input feature matrix $\textit{\textbf{M}}_{in}$, whose size is $a_{in} \times c_{in}$, only carries the information of all active points, and each row vector with the size of $1 \times c_{in}$ is the feature value of one active point across all $c_{in}$ channels. The hash table refers to the location map of active points in the input data grid, whose size is $h_{in} \times w_{in} \times l_{in}$, to the corresponding row index of the input feature matrix $\textit{\textbf{M}}_{in}$. The hash table is denoted as $\textit{\textbf{H}}_{in} = \{ (KEY_{i}, value_{key_{i}}): i\in\{0,1,2,....,(a_{in}-1)\} \} $, where $KEY_{i} = [x_{i}, y_{i}, z_{i}]$ is a $ 3\times 1$ vector of integer coordinates indicating the location of active points in the input data grid, and $value_{key_{i}}$ is the row index in $\textit{\textbf{M}}_{in}$.

The hash table can be built by iterating through all output points receiving information from any active input point. If the output point is visited for the first time, its spatial location and index is added to $\textit{\textbf{H}}_{out}$. The rule book, $\textit{\textbf{Rule}}$, depicting neuron connections from the current layer to the next layer is created based on the offset between input points and its corresponding output points. To keep output points in the center of the 3D convolutional kernel, an offset value is used to indicate the index of the input point in the 3D convolutional kernel. For example, the offset (0,0,0) indicates that the input point is located in the most top-right-front corner of the 3D kernel centered at one input, and the offset $(f-1,f-1,f-1)$ denotes the input point located in the most bottom-left-rear corner of the 3D kernel. Due to the intrinsic sparsity of the input data, the majority of points in the 3D kernel for one output point are frequently observed as inactive, thus an output point is usually only connected with a few active input points in a 3D kernel. The algorithm for generating the rule book, $\textit{\textbf{Rule}}$, the number of active points $a_{out}$, and hash table $\textit{\textbf{H}}_{out}$ for the next layer can be found in Appendix \ref{alg1}. 

The output feature matrix $\textit{\textbf{M}}_{out}$ with size $a_{out} \times c_{out}$ is computed by multiplying the convolutional weighting matrix $\textit{\textbf{W}}$ and input feature matrix. The size of matrix $\textit{\textbf{W}}$ is $c_{in}\times f^{3} \times c_{out}$ and $W^{(i,j,k)}$ is with size $c_{in}\times c_{out}$. This is the parameter weight for one offset or one connection between the input and output points. $\textit{\textbf{M}}_{out}$ is initialized with zeros. For one offset $(i,j,k)$, the feature matrix of all input points fetched according to $\textit{\textbf{Rule}}^{(i,j,k)}(:,0)$ is multiplied with the corresponding weighting parameters $W^{(i,j,k)}$, then scattered and added to the output feature matrix of the corresponding output points indexed by $\textit{\textbf{Rule}}^{(i,j,k)}(:,1)$, as shown in Fig. \ref{fig:sparseCNN}. After all offsets are determined, the bias vector $B$, whose size is $1 \times c_{out}$, is added to $\textit{\textbf{M}}_{out}$, followed by an activation function, Rectified Linear Unit (ReLU) \cite{Nair:2010:RLU:3104322.3104425}. 

Unlike a regular convolution, the sparse 3D CNN discards information from inactive points by assuming that the input from those points are zero rather than ground state. The algorithm for calculating the output feature matrix  $\textit{\textbf{M}}_{out}$ can be found in Appendix \ref{alg2} and Fig. \ref{fig:sparseCNN}. The hash table and Rule book generation will be implemented through Central Processing Unit (CPU) mode, and feature matrix computation will be implemented with a Graphical Processing Unit (GPU). The submanifold sparse CNN \cite{SubmanifoldSparseConvNet} restricts an output point to be active only if the point at the corresponding position in the input data is active. Therefore, the input hash table can be taken as the output hash table to increase the computation efficiency, and the sparsity of the 3D data remains the same.

\subsection{3D Backbone Network}
By using the sparse 3D CNN, this study aims to build 3D backbone networks for high resolution 3D feature maps with high semantic level for 3D object detection. The input for the 3D backbone network is voxelized into 3D grids, and outputs are 2D feature maps that are proportionally scaled at multiple levels. We propose two types of 3D backbone networks: 3DBN-1 and 3DBN-2, see Fig. \ref{fig:3DBN}.

{\bf 3DBN-1}: It is an unidirectional feed-forward computation of sparse 3D CNN and produces the pyramidal 3D feature hierarchy. Each pyramidal level consists of several 3D feature maps with the same size, and the Residual Network (ResNet) block is applied to process and generate these 3D feature maps. The sparse 3D CNN with the stride size $(2,2,2)$ is applied to reduce the 3D resolutions of feature maps and also increase the channel number $c_{i}$ to extract stronger semantic feature maps than the antecedents. At every pyramid level, a sparse 3D CNN kernel with size $(d_{i},1,1)$ is used to compress the 3D features into 2D feature maps for anchoring object proposals. The reason is that two 3D objects are rarely detected in the same vertical direction (Z-axis) in traffic scenes. We compress these 3D features into 2D maps to reduce the computation cost of generating 3D anchors because the 2D CNN is computationally less expensive than the sparse 3D CNN. There is usually a large semantic gap among different 3D feature maps of pyramidal levels in the unidirectional 3DBN-1, which is caused by various convolutional depths. The deeper the neural network, the stronger the semantic features \cite{Lin2016FeaturePN}. 

 {\bf 3DBN-2}: To close the semantic gap between the feature maps of different pyramidal levels, we build another 3D backbone network, 3DBN-2. Its bottom-top path is the same as 3DBN-1, however, it is added with an extra top-bottom path that passes the 3D semantic context downward. This may help to produce high-level 3D semantic feature maps at all pyramidal levels. The low-resolution but semantically strong feature maps are upsampled to the size of the upper pyramidal level by using the deconvolutional layers with the stride size of $(2,2,2)$, then concatenated to the corresponding feature maps of the same spatial size. The $3\times 3 \time 3$ sparse CNN is appended to the concatenated feature maps to diminish the effect of the upsampling and concatenation operation. Similarly, a sparse 3D CNN kernel with the size of $(d_{i},1,1)$ is applied to compress the 3D features into 2D feature maps; therefore, the size and type of its outputs are the same as that of 3DBN-1.

\section{3D Object Detection}\label{sec:3D_object_detection_network}

 The entire detection network consists of three main components: point cloud voxelization, 3D backbone network, and fusion network, as shown in Fig. \ref{fig:framework}. Voxelization converts raw point cloud into voxels which can be consumed by the 3D backbone. The fusion network generates 3D object proposals from feature maps provided by the 3D backbone network.

\subsection{Voxelization}
The input to the sparse 3D CNN should be in a regular grid, and the point cloud data is usually in an irregular format. Therefore, we cannot directly feed the point cloud into the aforementioned 3D backbone network. Voxelization is first designed to preprocess the unordered and irregular point cloud data to obtain its voxel representation. 

The 3D point cloud space range should be $l, w$, and $h$ along the $X$, $Y$, and $Z$ axes, respectively. The size of the voxel is set to $v_{l}, v_{w}$ and $v_{h}$. Therefore, the size of 3D voxel grid is ${l}/{v_{l}}$, ${w}/{v_{w}}$, and ${h}/{v_{h}}$. For the feature of every voxel, the simple Binary Voxel (BV) method is used to treat it as a binary value. The value 1 is assigned to the feature if there exists a point in corresponding voxel, otherwise, the voxel is empty and the feature value is 0. The drawback of this method is that the local shape information of points within a voxel is lost. To avoid the loss of local information, the voxel size in the BV is usually set very small to generate high-resolution 3D grids; for example, $v_{l}, v_{w}$ and $v_{h}$ at 0.025, 0.025, and 0.0375 meters are selected respectively in our experimental settings.

An alternative method is to use PointNet \cite{DBLP:journals/corr/QiSMG16} to extract the pointwise features, which is called Voxel Feature Encoding (VFE) \cite{DBLP:journals/corr/abs-1711-06396}. It enables inter-point interaction within a voxel by combining pointwise features with a locally aggregated feature, which can help to learn the local 3D shape information, and hence, can alleviate information loss due to data quantization. Since VFE requires a number of points per valid voxel, the voxel size in VFE should be large. Accordingly, we set $v_{l}, v_{w}$, and $v_{h}$ to 0.2, 0.2, and 0.3 meters, which are much larger than those in the BV. If this method is adopted, the voxelization would consist of two layers of VFE to learn the voxelwise feature with the size $1\times c_{0}$, and the shapes of the 3D voxel grids would remain the same. Therefore, ${h}/{v_{h}}, {w}/{v_{w}}$, and ${l}/{v_{l}}$ are set to $h_{0}, w_{0}$, and $l_{0}$, respectively.

\subsection{Fusion network}
\begin{figure}[!tbp]
	\includegraphics[width = \linewidth]{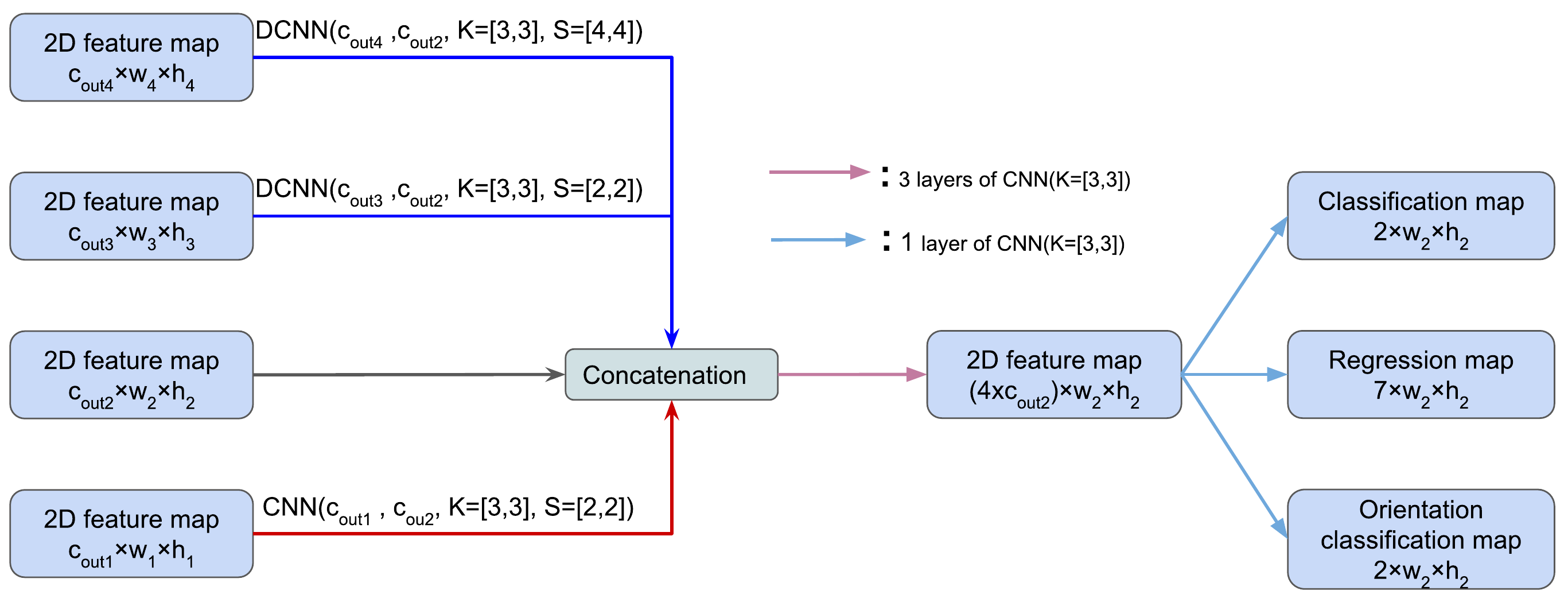} 
	\caption{Fusion Network: CNN is the 2D convolutional neural network, DCNN is the 2D deconvolutional neural network, and $K$ and $S$ are the convolutional kernel and stride, respectively.}
\label{fig:DN}
\end{figure}

The fusion network aims to detect objects from feature maps generated by the 3D backbone network, and it will fuse all the 2D feature maps together to form a final high resolution feature map with a high semantic level for classification and regression tasks. For object detection in the image, the scale of the object is unknown, and object detection is usually conducted on multiple feature map pyramids. High spatial resolution feature maps accommodate small objects and low spatial resolution feature maps are for large objects. However, the scale of 3D objects in the point cloud is almost fixed, and variations in 3D bounding box scales in the 3D space is much smaller than that of 2D bounding boxes in the image. Therefore, we concatenate all 2D feature maps, then employ three layers of conventional 2D CNN to fuse together feature maps of different spatial resolution. The classification, location, and orientation predication maps are generated on the final layer of the feature maps by using a layer of 2D CNN, see Fig. \ref{fig:DN}.

\subsection{Loss function}
The loss consists of three main components: classification, 3D bounding box regression, and heading classification, as given in Equation \ref{equ:all}.

\begin{equation}
L_{all}= \kappa L_{cls} + \lambda L_{reg} + \mu L_{ori}
\label{equ:all}
\end{equation}
where variables $\kappa$, $\lambda$ and $\mu$ are weighting factors to balance the relative importance.

\subsubsection
{Classification $L_{cls}$} The number of objects of interest in one frame is usually less than 30, while the number of assigned anchor boxes is around $ 140,000$. This leads to a significant imbalance between positive and negative labels. To address this, we calculate the classification loss for positive and negative anchors, separately. In addition, we applied the $\alpha-$balanced focal loss \cite{focalloss} for classification (see Equation \ref{equ:classification}). Variables $N_{pos}$ and $N_{neg}$ are the number of positive and negative labels; $p_{i}^{pos}$ and $p_{j}^{neg}$ are the softmax output from positive and negative anchors, respectively; $\alpha$ is the balanced parameters for positive and negative samples; and modulating factors $(1-p_{i}^{pos})^{\gamma}$ and $(p_{j}^{neg})^{\gamma}$ assign less weights and reduce the loss of well-classified examples.

\begin{equation}
\begin{aligned}
L_{cls}= & - \frac{1}{N_{pos}}\sum_{i} \alpha (1-p_{i}^{pos})^{\gamma} log(p_{i}^{pos})\\
&- \frac{1}{N_{neg}}\sum_{j}(1-\alpha)(p_{j}^{neg})^{\gamma}log(1-p_{j}^{neg})
\label{equ:classification}
\end{aligned}
\end{equation}

\subsubsection
{3D Bounding Box Regression $L_{reg}$} The regression targets ${\bf r^{*}}$ are defined as a $1\times7$ vector, $[r_{x}, r_{y}, r_{z}, r_{l}, r_{w}, r_{h}, r_{\theta}]$. Vector ${\bf r_{i}^{*}}$ encodes 3D ground truth bounding boxes based on anchor box correspondences, and the computation is defined in Appendix \ref{residual_com}. For the heading angle regression, Voxelnet \cite{DBLP:journals/corr/abs-1711-06396} uses angle residual between ground truth and anchor boxes directly, and AVOD-FPN \cite{avod_3d} encodes heading into the orientation vector by using trigonometric functions. Similarly, SECOND \cite{s18103337} employs sinusoidal functions to encode the radian offset. In contrast, this study proposes a simplified and effective angle loss function, which represents two opposite yaw angles in ground truth labels with an identical angle. The ground truth bounding boxes with opposite headings are treated identically, thus, the original range of heading, $2\pi$, will be changed into $\pi$, and discontinuities of angle residual loss value caused by two opposite headings of the same ground truth bounding boxes will be mitigated. The orientation information is abandoned in 3D bounding box regression but is considered in heading classification. Hence, only small and continuous angle offsets $r_{\theta}$ with regard to anchor boxes are encoded. The $smoothL1$ function \cite{smoothl1} is used to calculate the loss value, as given in Equation \ref{equ:regression}.  

\begin{equation}
\begin{aligned}
L_{reg}=& \frac{1}{N_{pos}}\sum_{i} SmoothL1(\bf {r_{i}^{*}},\bf{r_{i}})
\label{equ:regression}
\end{aligned}
\end{equation}

\subsubsection
{Heading Classification $L_{ori}$} To retain the orientation of objects, we adopt an extra softmax loss function, as given in Equation \ref{equ:orientation}. 

\begin{equation}
\begin{aligned}
L_{ori}=& -\frac{1}{N_{pos}}(\sum_{i}log(p_{i}^{+pos}) + \sum_{j}log(1-p_{j}^{-neg}) )
\label{equ:orientation}
\end{aligned}
\end{equation}
The ground truth bounding boxes whose heading angles are larger than zero, are trained as positive labels, and the rest are regarded as negative labels. The variables $p_{i}^{+pos}$ and $p_{j}^{-neg}$ are the softmax output for positive and negative labels, respectively.

\section{Experiments}\label{sec:experiments}

\subsection{Implementation details}
\subsubsection{Dataset}
The KITTI benchmark dataset \cite{geiger2012we} is adopted to validate the proposed methods. It includes 7481 training and 7518 testing sets of LIDAR laser data. Since the ground truth of the testing dataset is not publicly available, we split the training dataset in a 1:1 ratio for training and validation, then used them to conduct comparative experiments for tuning the models and comparing other methods. The adequately trained network model with the optimal configuration is deployed to process the testing dataset, and their results are submitted to the KITTI benchmark.

\begin{figure*}[ht]
	\includegraphics[ width = \linewidth]{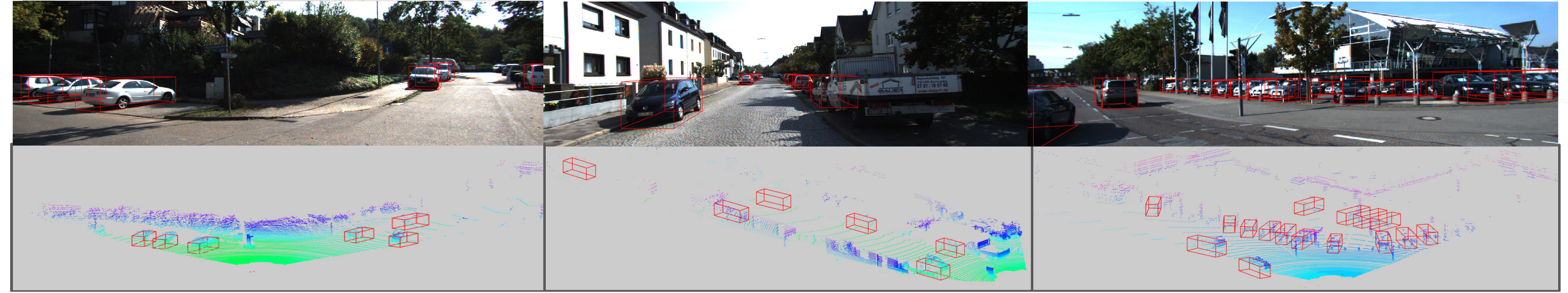} 
	\caption{Visualization of 3D object detection on the KITTI testing dataset. Top row: 3D bounding boxes in RGB images. Bottom row: corresponding bounding boxes in point cloud.}
\label{fig:sample_img}
\end{figure*}

\begin{table*}[ht]
\caption{Comparison of 3D detection models with different combinations of voxelization and 3D backbone networks on validation dataset. }
\label{table:tunning_model}
\centering
	\begin{tabular}{c|c|c|c|c|c|c}
	\hline
	Voxelization & Backbone Network & Easy & Moderate & Hard & mAP & Time(s) \\ \hline
    BV & 3DBN-1 & 87.6 & 75.34 & 75.02  & 79.32 & 0.12  \\ \hline
	VFE & 3DBN-1 & 86.81 & 76.12 & 74.19  & 79.04 & {\bf0.058} \\ \hline
	BV & 3DBN-2 & 87.98 & {\bf77.89} & {\bf76.35} & {\bf80.74} & 0.13  \\ \hline
	VFE & 3DBN-2 & {\bf88.20} & 77.59 & 75.58 & 80.46 & 0.068 \\ \hline
	\multicolumn{7}{p{10cm}}{mAP(\%): average precision of Easy, Moderate and Hard categories for 3D car detection, time(s): time needed for processing one sample.}
	\end{tabular}
\end{table*}

\begin{table*}[ht]
\caption{Comparison of the proposed methods with others for 3D car detection on validation datasset.}
\label{table:comparsion_methods}
\centering
	\begin{tabular}{l|c|c|c|c|c|c}
	\hline
	Methods & Modality & Easy & Moderate & Hard & mAP & Time(s)\\ \hline
	BV+3DBN-2& LiDAR & 87.98 & \bf{77.89} & \bf{76.35} & \bf{80.74} & 0.13 \\ \hline
	VEF+3DBN-2& LiDAR & \bf{88.20} & 77.59 & 75.58 & 80.46 & 0.068 \\ \hline
	SECOND\cite{s18103337}& LiDAR & 87.43 & 76.48 & 69.10 & 77.67 & \bf{0.05} \\ \hline
	AVOD-FPN\cite{avod_3d}& LiDAR + Camera & 84.41 & 74.44 & 68.65 & 75.83 & 0.1 \\ \hline
	VoxelNet\cite{DBLP:journals/corr/abs-1711-06396} & LiDAR & 81.97  & 65.46 & 62.85 & 70.09 & 0.23 \\ \hline
	F-PointNet\cite{DBLP:journals/corr/abs-1711-08488} &LiDAR + Camera & 83.76  & 70.92 & 63.65 & 72.78 & 0.36 \\ \hline	
    MV3D\cite{DBLP:journals/corr/ChenMWLX16} &LiDAR + Camera & 71.29  & 62.68 & 56.56 & 63.51 & 0.17 \\ \hline
    \multicolumn{7}{p{10cm}}{The performance is illustrated by AP(\%) of every category.}
	\end{tabular}
\end{table*}

\subsubsection{Network details}
The ranges of point cloud considered were [0, 70.2], [-39.9, 39.9] and [-3.25, 1.25] meters along the $X$-, $Y$-, and $Z$-axis, respectively. If BV is selected for voxelization, the 3D voxel size is [0.025, 0.025, 0.0375] meters along the $X$-, $Y$-, and $Z$-axis. The voxelizaiton consists of four layers of sparse 3D CNN block with a convolutional kernel size [3,3,3] and
a stride size [2,2,2]. Sizes of the corresponding feature maps are $16(c)\times127(h)\times3199(w)\times2815(l)$, $32\times63\times1599\times1407$, $48\times31\times799\times703$, and $64\times15\times399\times351$. The parameters $[c_{0},h_{0},w_{0}, l_{0}]$ are set to $[64, 15, 399, 351]$. When VEF is used, the voxelization includes two layers of VEF to increase the number of feature channels from 7 to 32 and from 32 to 128, and $c_{0}$ is 128. The 3D voxel sizes are [0.2, 0.2, 0.3] meters and $[h_{0},w_{0}, l_{0}]$ are $[15, 399, 351]$. The maximum number of kept points per voxel is 35. As for the 3D backbone networks and the fusion network, $[c_{1}, c_{2}, c_{3}, c_{4}]$ are set to $[64, 80, 96, 128]$; $[h_{1}, h_{2}, h_{3}, h_{4}]$ are $[15, 7, 3, 1]$; $[w_{1}, w_{2}, w_{3}, w_{4}]$ are $[399, 199, 99, 49]$; and $[l_{1}, l_{2}, l_{3}, l_{4}]$ are $[351, 175, 87, 43]$. Dimensions of the 2D feature maps $[c_{out1}, c_{out2}, c_{out3}, c_{out4}]$ are $[128, 128, 128, 128]$.

The first set of anchors contains 3D cube boxes with size [1.6 (l), 3.9 (w), 1.56 (h)] meters and orientation [0, $\pi/2$]. The second set of anchors contains 3D cube boxes with size [1.581, 3.513, 1.511] and [1.653, 4.234, 1.546] meters and orientation [0, $\pi/2$]. Two sets of sizes are obtained by using K-means clustering, with K equal to 2, among all 3D ground truth bounding boxes. Anchors whose Intersection over Union (IoU) with the ground truth are over 0.7 for the car are treated as a positive label, and those IoU less than 0.5 for the car are regarded as a negative label. Coefficients $\alpha$, $\beta$, and $\delta$ used in the loss function are 1, 2, and 1, respectively.

The AdamOptimizer \cite{DBLP:journals/corr/KingmaB14} was employed to train our network, and it was configured with an initial learning rate of 0.0002 and an exponential decay factor of 0.8 for every 18570 steps.

\subsubsection{Data augmentation}
To reduce the overfitting caused by the fact that the complex model is trained from scratch on the small dataset, four techniques of data augmentation introduced in this method are:

\paragraph
{Rotation} every point cloud in the entire block is rotated along the Z-axis and about the origin [0,0,0] by $\theta$, and $\theta$ is drawn from the uniform distribution [-$\pi$/4, $\pi$/4]. 

\paragraph
{Scaling} similar to the global scaling operation in image-based data augmentation, the whole point cloud block is zoomed in or out. The $XYZ$ coordinates of points in the whole block are multiplied with a random variable which is drawn from the uniform distribution [0.95, 1.05].

\paragraph
{Motion} in order to increase the variation of each ground truth 3D bounding box, a motion process is applied to independently perturb each bounding box. The 3D bounding box is first rotated around the Z-axis by a uniformly distributed random variable, $\delta \in [\pi/2, \pi/2]$. Then, a translation $[\triangle X, \triangle Y, \triangle Z]$ is used to move the rotated bounding box, and  $[\triangle X, \triangle Y, \triangle Z]$ are independently sampled from a zero mean and unit standard deviation Gaussian distribution. The collision test will be conducted after one motion.

\paragraph
{Increment} there are usually only a few ground truth 3D bounding boxes in one point cloud block, which leads to a very small number of positive samples. To diversify training scenarios and increase the number of positive samples, we insert some extra 3D ground truth bounding boxes and their corresponding points into the current training samples. Extra 3D bounding boxes are randomly picked from the database, which includes all ground truth bounding boxes in the training dataset. Similarly, the collision test is conducted after the increment. 

\subsection{Comparisons on the KITTI validation dataset}

Two kinds of 3D backbone network architectures and two different voxelization techniques were applied to generate 3D voxel grid input. To find the performance of various combinations, we conducted experiments on the KITTI validation dataset. The result can be found in Table \ref{table:tunning_model}, and shows that the combination of BV and 3DBN-2 achieves the best performance. However, the BV voxelization led to a large increase in the processing time. The main reason is that the high-resolution 3D voxel grid requires more operations in the rule book generation，and it is the most time-consuming procedure in the sparse CNN. On the other hand, the VFE is faster because the PointNet computation mainly consists of one dimensional CNN and max pooling, and it can be efficiently implemented on a GPU. In addition, 3DBN-2, which is designed to generate a better 3D semantic feature map than 3DBN-1, shows a better performance than 3DBN-1 with little extra computation.  

We also conducted various experiments on the validation dataset to compare our 3D object detection methods. The combination of BV and 3DBN-2 and that of VEF and 3DBN-2 were selected to perform comparisons with other methods. Comparison results are shown in Table \ref{table:comparsion_methods}, from which we can find the combination of VEF and 3DBN-2 achieves the best performance for the Easy category, and the combination of BV and 3DBN-2 is the best for the Moderate and Hard categories. The MV3D, which converted the point cloud into image representations and applied 2D CNN for 3D object detection, has the worse accuracy than other detection methods based on point cloud. We can conclude that 3D feature maps are better for 3D object detection than 2D feature maps extracted from bird view point cloud. This is also supported by the experimental results that our 3D backbone network based detection methods achieve the best detection accuracy. In addition, unlike the AVOD-FPN, F-PointNet, and MV3D using both LiDAR and camera data, our method solely relies on the LiDAR.

\subsection{Performance on the KITTI testing dataset}
The labeled KITTI testing dataset is not released for the public, and detection performance can only be evaluated in the KITTI server after participants submit detection results from the testing dataset to the server. We chose the combination of BV and 3DBN-2 on KITTI to process the testing dataset, as this combination can produce the best performance in the validation dataset. The performance is shown in Tables \ref{table:testing_dect} and \ref{table:testing_ori}, and several samples are displayed in Fig. \ref{fig:sample_img}. When we submitted the results of the testing dataset to KITTI evaluation server on 23 January 2019, our method ranked top three among all known methods with released publications on the KITTI benchmark\footnote[3]{Testing results can be found in the KITTI object detection leaderboard \url{http://www.cvlibs.net/datasets/kitti/eval_object.php?obj_benchmark=3d}.}. This indicates that our method can reach top performance in comparison with other 3D object detection methods. 

\begin{table}[ht]
\caption{Comparison of proposed methods with others on KITTI testing dataset for 3D car detection}
\label{table:testing_dect}
\centering
	\begin{tabular}{c|c|c|c|c}
	\hline
	Methods & Modality & Easy & Moderate & Hard \\ \hline
	Our \bf{3DBN}& LiDAR &\bf{83.56} & \bf{74.64} & \bf{66.76} \\ \hline
	SECOND\cite{s18103337}& LiDAR & 83.13 & 73.66 & 66.20  \\ \hline
	AVOD-FPN\cite{avod_3d}& LiDAR + Camera & 81.94 & 71.88 & 66.38  \\ \hline
	VoxelNet\cite{DBLP:journals/corr/abs-1711-06396}& LiDAR & 77.47  & 65.11 & 57.73  \\ \hline
	F-PointNet\cite{DBLP:journals/corr/abs-1711-08488}& LiDAR + Camera & 81.20  & 70.39 & 62.19  \\ \hline	
    MV3D\cite{DBLP:journals/corr/ChenMWLX16}& LiDAR + Camera & 71.09  & 62.35 & 55.12  \\ \hline
    \multicolumn{5}{p{7.5cm}}{The accuracy data is available in KITTI benchmark online ranking$^{[3]}$.}
	\end{tabular}
\end{table}

\begin{table}[ht]
\caption{3D car orientation prediction performance of our proposed method in comparison with others on KITTI testing dataset.}
\label{table:testing_ori}
\centering
	\begin{tabular}{c|c|c|c|c}
	\hline
	Methods & Modality & Easy & Moderate & Hard \\ \hline
	Our \bf{3DBN}& LiDAR & 89.93 & \bf{87.95} & 79.32 \\ \hline
	SECOND\cite{s18103337}& LiDAR & 87.84 & 81.31 & 71.95  \\ \hline
	AVOD-FPN\cite{avod_3d} & LiDAR + Camera & \bf{89.95} & 87.13 & \bf{79.74}  \\ \hline
    \multicolumn{5}{p{7.5cm}}{The accuracy data is available in KITTI benchmark online ranking$^{[3]}$.}
	\end{tabular}
\end{table}


Table \ref{table:testing_dect} shows the 3D object detection accuracy which can be found on the KITTI benchmark webpage, and the performance is very similar with that on the validation dataset, see Table \ref{table:comparsion_methods}. The MV3D is ranked at the bottom and our proposed 3D backbone network achieves the top performance compared with other popular 3D detection methods. This further supports our hypothesis that 3D feature maps are better for 3D object detection tasks than other methods that compress 3D data into 2D \cite{DBLP:journals/corr/ChenMWLX16, avod_3d} or employ a shallow 3D network as a supplement for a 2D region proposal network \cite{DBLP:journals/corr/abs-1711-06396, s18103337}. 

The orientation prediction task is compared in Table \ref{table:testing_ori}, yet only includes AVOD-FPN\cite{avod_3d} and SECOND\cite{s18103337} due to the missing orientation results of other methods. Our simple and efficient orientation regression method can achieve competitive performance with AVOD-FPN, which employs complex encoding mechanism.  

\section{Conclusion}\label{sec:conclusion}
Based on the observation that 2D backbone networks are vital to the  success  of  2D  CNN  based object  detection in computer vision, we propose the 3D backbone network for 3D object detection in point cloud. The 3D backbone network is made up of a sparse 3D CNN and can be implemented efficiently. Experimental results on the KITTI benchmark shows that our proposed detection method using 3D backbone network overpasses other approaches that convert 3D data into 2D format and apply 2D CNN to process data. Our 3D backbone network can extract better 3D feature maps than others for 3D object detection. To extract richer feature information than point cloud based methods, further research work will be focused on extending the 3D backbone network to fuse the image and point cloud together to improve object detection performance.

\section{Acknowledgment}
The presented research is supported by the China Scholarship Council (Grant no. 201606950019).

\balance

\bibliographystyle{IEEEtran}
\bibliography{IEEEabrv,Example}

\appendices
\vspace{30 mm}

\section{}\label{alg1}
\begin{algorithm}[H]\footnotesize
\caption{Ruler book and output hash table generation}
\label{alg:1}
\begin{algorithmic}
\STATE Legend:
\STATE
\begin{tabular}{p{0.5cm}p{7cm}}
$H_{in}$:  & hash table of input, active points coordinates: $key_{i} = (x_{i}, y_{i}, z_{i})$\\
$H_{out}$: & hash table of output data\\
$Cntr$: & number of input and output pairs in the rule book\\
$a_{in}$: & number of active point in input data\\
$f$: & size of 3D CNN kernel\\
$s$: & stride in 3D CNN\\
$GetOutputCoord$: \\& get all output points in regards to one input point\\
$GetOffset$:\\ & calcuate offset of input point with regard to output point in the 3D convolutional kernel\\
\hline
\end{tabular}
\STATE $H_{out} \leftarrow [ ] $; $Rule \leftarrow [] $; $Cntr \leftarrow [] $; $a_{out} \leftarrow 0 $
\FOR{$i_{in} \leftarrow 0$ to $a_{in}$}
\STATE $P_{in}\leftarrow key_{i_{in}}$ in $H_{in}$
\STATE $\textit{\textbf{P}}_{out} \leftarrow GetOutputCoord(P_{in}, k, s)$
        \FOR{$P \in \textit{\textbf{P}}_{out}$}
            \IF{$P \notin$ $key$ in $H_{out}$}
            \STATE $key_{a_{out}} \leftarrow P$; $value_{key_{a_{out}}} \leftarrow a_{out}$; $a_{out} \leftarrow a_{out}+1$
            \ENDIF
            \STATE $(i,j,k) \leftarrow GetOffset(P_{in}, P)$
            \STATE $\textit{\textbf{Rule}}^{(i,j,k)}(Cntr(i,j,k), 0) \leftarrow i_{in}$
            \STATE $key \leftarrow P$
            \STATE $\textit{\textbf{Rule}}^{(i,j,k)}(Cntr(i,j,k), 1) \leftarrow value_{key}$
            \STATE $Cntr(i,j,k) \leftarrow Cntr(i,j,k)+1$
        \ENDFOR
\ENDFOR
\end{algorithmic}
\end{algorithm}
\section{}\label{alg2}
\begin{algorithm}[H]\footnotesize
\caption{Output feature matrix generation}
\label{alg:2}
\begin{algorithmic}
\STATE Legend:
\STATE
\begin{tabular}{p{0.5cm}p{7cm}}
$f$: & size of the 3D CNN kernel\\
$B$: & bias vector of 3D CNN with size $1 \times n_{out}$\\
$W$: & the 3D convolutional parameter matrix with the size,  $n_{in}\times f^{3} \times n_{out}$, $W^{(i,j,k)}$ is the parameter weight for one offset or connection with size $n_{in}\times n_{out}$ \\
$ActivationFunc$:\\ & activation function used by the sparse 3D CNN\\
\hline
\end{tabular}
\STATE $\textit{\textbf{M}}_{out}$: output feature matrix
\STATE $\textit{\textbf{M}}_{out} \leftarrow 0 $
\FOR{ $i \leftarrow 0$ to $f$}
    \FOR{ $j \leftarrow 0$ to $f$}
        \FOR{ $k \leftarrow 0$ to $f$}
            \FOR{$idx \leftarrow 0$ to $Cntr(i,j,k)$}
                \STATE $i_{in} \leftarrow \textit{\textbf{Rule}}^{(i,j,k)}(idx,0)$
                \STATE $i_{out} \leftarrow \textit{\textbf{Rule}}^{(i,j,k)}(idx,1)$
                \STATE $\textit{\textbf{M}}_{out}(i_{out},:) \leftarrow \textit{\textbf{M}}_{out}(i_{out},:)+\textit{\textbf{M}}_{in}(i_{in},:)\times W^{(i,j,k)}$
            \ENDFOR
        \ENDFOR    
    \ENDFOR
\ENDFOR
\STATE $\textit{\textbf{M}}_{out} \leftarrow \textit{\textbf{M}}_{out} + B$
\STATE $\textit{\textbf{M}}_{out} \leftarrow ActivationFunc(\textit{\textbf{M}}_{out})$
\end{algorithmic}
\end{algorithm}

\section{}\label{residual_com}
\begin{footnotesize}
\textbf{Residual vector computation}: the 3D ground truth bounding box is parameterized as $[x_{gt}, y_{gt}, z_{gt}, l_{gt}, w_{gt}, h_{gt}, \theta_{gt}]$, and its corresponding anchor boxes are $[x_{a}, y_{a}, z_{a}, l_{a}, w_{a}, h_{a}, \theta_{a}]$, where $[x, y, z]$ denote the center locations, $[l, w, h]$ are the length, width, and height of the 3D bounding box, and $\theta$ is the yaw angle around the Z-axis.

${\bf r^{*}} = [r_{x}, r_{y}, r_{z}, r_{l}, r_{w}, r_{h}, r_{\theta}]$ can be computed as following:

\begin{equation}
\label{equ::residul_com}
\begin{aligned}
& r_{x} = \frac{x_{gt} - x_{a}}{d_{a}}, r_{y} = \frac{y_{gt} - y_{a}}{d_{a}}, r_{z} = \frac{z_{gt} - z_{a}}{h_{a}}\\
& r_{l} = log(\frac{l_{gt}}{l_{a}}), r_{w} = log(\frac{w_{gt}}{w_{a}}), r_{h} = log(\frac{h_{gt}}{h_{a}})\\
& u_{\theta} = \theta_{gt} - \theta_{a}, d_{a} = \sqrt{l_{a}^x + w_{a}^2}
\end{aligned}
\end{equation}
\end{footnotesize}

\begin{IEEEbiography}[{\includegraphics[width=1in,height=1.25in,clip,keepaspectratio]{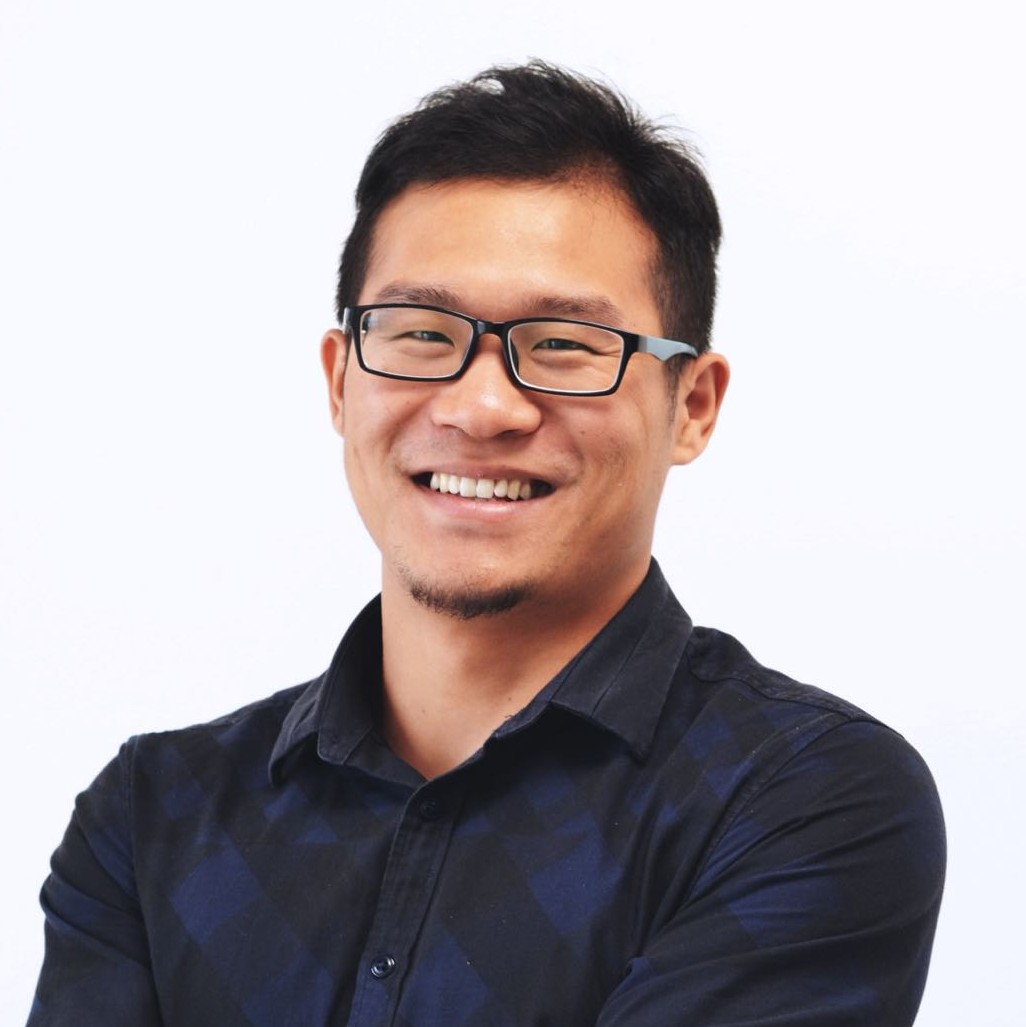}}]{Xuesong LI}
received his M.Sc. degree in Automotive Engineering from Wuhan University of Technology, Hubei, China, in 2016. He is currently a Ph.D. student in Mechatronics, at the School of Mechanical Engineering, University of New South Wales, Sydney. His research topic is about perception system for autonomous driving vehicles.
\end{IEEEbiography}
\vskip -2\baselineskip plus -1fil
\begin{IEEEbiography}[{\includegraphics[width=1in,height=1.25in,clip,keepaspectratio]{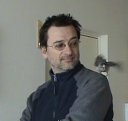}}]{Jose E Guivant}
obtained his Ph.D. degree, in Robotics, from The University of Sydney, Australia, in July/2002. He is currently Sr. Lecturer in Mechatronics, at the School of Mechanical Engineering, University of New South Wales, Australia.
\end{IEEEbiography}
\vskip -2\baselineskip plus -1fil
\begin{IEEEbiography}[{\includegraphics[width=1in,height=1.25in,clip,keepaspectratio]{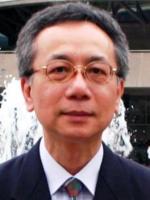}}]{Ngai Ming Kwok}
received the Ph.D. degree in mobile robotics from the University of Technology, Sydney, Australia, in 2007. He is currently a Lecturer with the School of Mechanical and Manufacturing Engineering, University of New South Wales, Sydney. His research interests include intelligent computation, image processing, and automatic control.
\end{IEEEbiography}
\vskip -2\baselineskip plus -1fil
\begin{IEEEbiography}[{\includegraphics[width=1in,height=1.25in,clip,keepaspectratio]{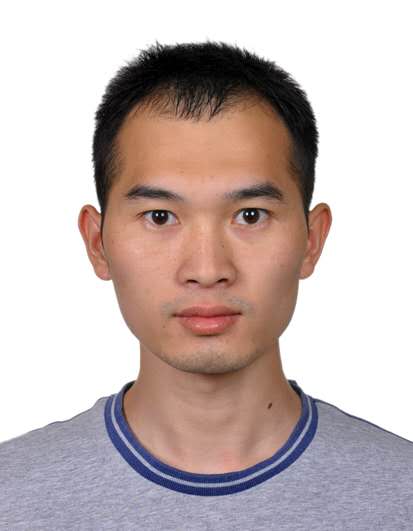}}]{Yongzhi Xu}
received his M.E. degree in Precious Instruments and Machinery from Beihang University in 2015. He is currently a Ph.D. student at the school of Civil and Environmental Engineering, University of New South Wales, Australia.
\end{IEEEbiography}

\vskip -2\baselineskip plus -1fil
\begin{IEEEbiography}[{\includegraphics[width=1in,height=1.25in,clip,keepaspectratio]{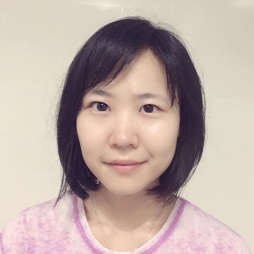}}]{Ruowei Li}
received her B.E. in Mechatronics Engineering from the University of New South Wales, Australia, where she is  currently working toward the PhD degree. Her research interests include computer vision, image processing and machine learning. 
\end{IEEEbiography}

\vskip -2\baselineskip plus -1fil
\begin{IEEEbiography}[{\includegraphics[width=1in,height=1.25in,clip,keepaspectratio]{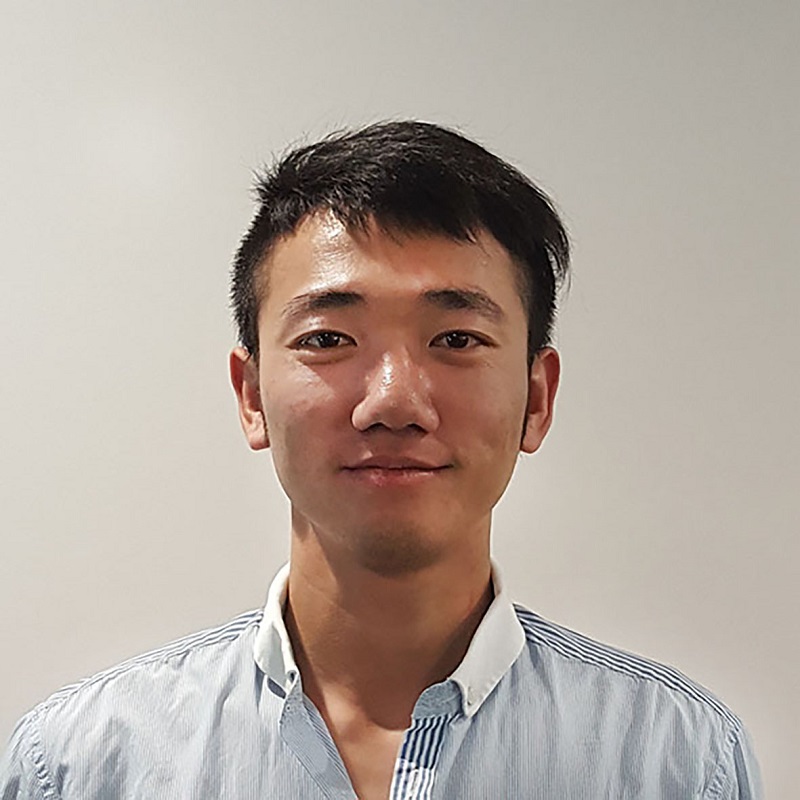}}]{Hongkun Wu}
Obtained his M.E. degree in Mechanical Engineering from Xi'an Jiaotong University, Xi'an, China, in 2015. He is currently a Ph.D. student at the school of Mechanical and Manufacturing Engineering, University of New South Wales, Australia. His research interests include machine condition monitoring and image processing.
\end{IEEEbiography}

\end{document}